\documentclass[journal]{IEEEtran}
\usepackage[T1]{fontenc}
\usepackage[utf8]{inputenc}
\usepackage{amsmath,amsfonts,amssymb}
\usepackage{mathrsfs}
\usepackage{array}
\usepackage{booktabs}
\usepackage{cite}
\usepackage{graphicx}
\usepackage{stfloats}
\usepackage{tabularx}
\usepackage{textcomp}
\usepackage{url}
\newcolumntype{Y}{>{\raggedright\arraybackslash}X}
\graphicspath{{pic/}}
\begin{document}

\title{One-Step Evolution for Long-Time Extrapolation:\\
An Error-Bound-Informed and Prior-Guided Neural Residual Framework for Autonomous PDEs}

\author{Maqun Zhang$^1$, Feng Gao$^1$, Wankun Chen$^1$, Hui Yu$^2$, Yanhai Gan$^{1,*}$, and Junyu Dong$^{1,*}$%
\thanks{This work was supported by the Leverhulme Trust through Project VP1-2020-044; the National Natural Science Foundation of China under Grant 42406192; the Fundamental Research Funds for the Central Universities under Grants 202413040 and 202572015; the National Science and Technology Major Project of China under Grant 2022ZD0117201; and the Postdoctoral Project of Qingdao under Grant QDBSH20240102021.}%
\thanks{Corresponding authors: Yanhai Gan (ganyanhai@ouc.edu.cn) and Junyu Dong (dongjunyu@ouc.edu.cn).}%
\thanks{Maqun Zhang, Feng Gao, Wankun Chen, Yanhai Gan, and Junyu Dong are with the State Key Laboratory of Physical Oceanography and Faculty of Information Science and Engineering, Ocean University of China, Qingdao 266100, China. Hui Yu is with the School of Psychology and Neuroscience, University of Glasgow, Glasgow G12 8QQ, U.K.}}

\maketitle

\begin{abstract}
Accurate simulation of the long-time evolution of systems governed by partial differential equations (PDEs) is central to scientific computing. Among existing deep learning-based approaches for solving PDEs, neural operators typically rely on extensive trajectory data, whereas physics-informed methods often exhibit limited stability during long-time extrapolation. For a well-posed autonomous PDE, long-time trajectories can be generated by repeated composition of a fixed-step evolution operator; hence, long-time extrapolation depends on controlling the approximation error of this operator and the propagation of that error under recursive composition. Accordingly, we propose a numerical-prior-guided, physics-constrained method trained without ground-truth trajectory supervision: a low-cost numerical prior reduces the difficulty of approximating the one-step evolution operator, while a weak-form PDE residual provides a computable proxy for the one-step error term in the error-propagation bound. We validate the method on five benchmark cases spanning four PDE classes and compare it with ten physics-informed learning methods under a unified protocol that excludes ground-truth trajectories from training and model selection. The results indicate that, in all five cases, the proposed method reduces long-time extrapolation error relative to the numerical prior and outperforms the best competing baseline in each case, thereby improving long-time simulation accuracy across different PDEs without ground-truth trajectory supervision. The source code developed for this paper will be made publicly available upon acceptance of the manuscript.
\end{abstract}


\section{INTRODUCTION}

Partial differential equations (PDEs) are widely used to describe spatiotemporal processes, including fluid motion, heat transfer, wave propagation, and phase-field evolution. Accurate simulation of the long-time evolution of these processes from prescribed initial conditions is fundamental to scientific computing and prediction in engineering \cite{ref1,ref2}. For a well-posed autonomous PDE with specified physical parameters and boundary conditions, the state transition over a fixed time interval can be represented by a well-defined evolution operator \cite{ref3,ref4}. Deep neural networks can approximate complex nonlinear mappings and continuous operators, thereby providing a new means of learning this evolution operator \cite{ref5,ref6}.

Deep learning-based PDE solvers primarily follow two paradigms: data-driven neural operators and physics-informed methods \cite{ref7,ref8,ref9}. Neural operators such as DeepONet and the Fourier neural operator (FNO) learn complex function-to-function mappings but typically require paired samples generated by high-fidelity solvers. For time-dependent problems, these samples take the form of trajectories, and generalization error depends on the input-function distribution and training-sample coverage \cite{ref7,ref8,ref10}. Physics-informed neural networks (PINNs) and their variants reduce reliance on ground-truth data by incorporating governing equations and initial and boundary conditions \cite{ref9}. However, conventional PINNs directly optimize the full spatiotemporal solution and remain subject to loss ill-conditioning, gradient imbalance, and temporal-causality difficulties for complex equations and long time intervals \cite{ref11,ref12,ref13}. Physics-constrained autoregressive networks and PhyCRNet can recursively solve time-dependent PDEs without ground-truth trajectory supervision \cite{ref14,ref15}, but one-step errors accumulate during rollout, limiting the transfer of training-window accuracy to long-time extrapolation \cite{ref16,ref17}.

Numerical methods can provide physically meaningful state updates for deep learning models \cite{ref18,ref19,ref20}, but finite resolution still introduces discretization errors \cite{ref21,ref22}. Existing numerical-learning hybrid methods often rely on high-fidelity trajectory supervision or corrections tailored to specific numerical schemes and are therefore not readily applicable to long-time extrapolation without ground-truth trajectories \cite{ref18,ref19,ref20}. Accordingly, this work investigates how a numerical prior can reduce the difficulty of approximating the evolution operator and constrain its approximation error without ground-truth trajectory supervision, allowing a model trained within a short time window to be applied to long-time extrapolation.

For an autonomous PDE that is well-posed over the target time horizon, the solution flow forms a time-homogeneous semigroup; consequently, long-time evolution can be represented by repeated composition of the same fixed-step state-transition operator \cite{ref3,ref4}. This structure allows the one-step dynamics learned within a short time window to be reused at later times, but errors introduced by repeated application of the approximate operator continue to propagate; therefore, the recursive structure of the exact evolution does not automatically translate into long-time model accuracy \cite{ref16,ref17}. Over the set of states reachable during extrapolation, long-time error is jointly controlled by the uniform approximation error of the one-step operator and system stability. Accordingly, using an existing physical approximation and correcting its one-step error, rather than learning the full evolution operator, provides a natural means of reducing approximation difficulty and improving long-time extrapolation.

To realize this one-step error correction, the numerical prior first produces a state prediction, while a fully convolutional network learns a correction; the next state is obtained through additive fusion and hard enforcement of physical constraints. Because the ideal correction is unavailable, we construct a physics-based loss from the weak-form PDE residual \cite{ref23} and, based on Eq. (7), use it as a conditional a posteriori proxy for the one-step approximation error of the evolution operator, thereby training the correction network without ground-truth trajectory supervision. The model is trained only within a short time window and then applied recursively beyond the training interval.

The main contributions are summarized as follows:

\begin{enumerate}

\item \textbf{An operator-approximation framework for long-time extrapolation with neural PDE solvers.} Starting from the existence and recursive structure of the fixed-step evolution operator for a well-posed autonomous PDE, we apply the one-step error-propagation relation to neural approximation models, identify the one-step approximation error as the key quantity linking short-time training to long-time extrapolation, and establish the design principles for numerical-prior guidance and physics-based constraints without ground-truth trajectory supervision.

\item \textbf{A prior-correction and error-constraint mechanism without ground-truth trajectory supervision.} A low-cost numerical prior provides the baseline evolution update, a correction network corrects errors in the prior update, and the weak-form PDE residual serves as a conditional a posteriori proxy for the one-step error; hard constraints preserve the initial and boundary conditions.

\item \textbf{A unified evaluation across PDEs and solution structures.} Under a unified protocol that excludes ground-truth trajectories from training and model selection, we compare the proposed method with ten physics-informed learning methods on five benchmark cases spanning four PDE classes and systematically evaluate its long-time extrapolation performance relative to the numerical prior and the competing methods.

\end{enumerate}

\section{RELATED WORK}

\subsection{Data-Driven Neural Operators}

Neural operators directly learn function-space mappings from initial and boundary conditions, equation coefficients, or source terms to PDE solutions. DeepONet, FNO, and neural operator theory form the foundations of this line of research \cite{ref6,ref7,ref8}. Recent surveys have organized the broader landscape of deep neural networks for PDEs, including physics-informed solvers and neural operators \cite{ref42}. Subsequent studies incorporated PDE constraints through PINO \cite{ref24} and extended neural operators to complex geometries \cite{ref25}, cross-equation pretraining \cite{ref26,ref27}, and time-invariant evolution modeling \cite{ref28}. Recent graph neural operator architectures further exploit multiscale spatial and frequency-domain features for PDE solving and retain strong performance with limited training samples and low-resolution data \cite{ref43}. Although these methods improve operator representation and transferability, most still rely on paired solution data or large-scale pretraining trajectories, limiting their long-time extrapolation by data costs and training-distribution coverage.

\subsection{Physics-Informed PDE Solvers}

PINNs incorporate governing equations and initial and boundary conditions into the loss function, thereby directly approximating PDE solutions without ground-truth data \cite{ref9}; however, multi-term losses involving differential operators often lead to gradient imbalance and ill-conditioned optimization \cite{ref11,ref12}. To address these issues, VPINN \cite{ref29} and hp-VPINN \cite{ref23} introduce weak-form constraints; SA-PINN \cite{ref30}, gPINN \cite{ref31}, and Causal PINN \cite{ref13} improve training through loss weighting, gradient enhancement, and temporal causality, respectively; and PINNsFormer \cite{ref32} and RoPINN \cite{ref33} further improve network architectures and sampling strategies. Benchmark studies indicate that these variants still exhibit inconsistent performance on complex PDEs \cite{ref34}; moreover, most methods directly optimize the full solution over a prescribed spatiotemporal domain, so extrapolation beyond the temporal domain is not a natural output of their formulations.

\subsection{Autoregressive Neural PDE Solvers}

Autoregressive methods learn fixed-step state-transition mappings and generate long-time trajectories by repeatedly applying the same model. PDE-Net was among the early methods to employ constrained convolutions for dynamical time stepping \cite{ref35}. AR-DenseED and PhyCRNet further combined physical constraints with convolutional recurrent architectures to enable recursive PDE solution without ground-truth trajectory supervision \cite{ref14,ref15}, while graph-network methods extended this paradigm to irregular discretizations \cite{ref16,ref17}. The main challenges are train--inference distribution shift and the accumulation of one-step errors; PDE-Refiner, APEBench, and recurrent neural operators investigate long-rollout stability through prediction refinement, unified benchmarking, and recurrent training, respectively \cite{ref36,ref37,ref38}. However, most methods still rely on ground-truth trajectories, while the long-time accuracy of physics-constrained models trained without ground-truth trajectory supervision remains controlled by the one-step approximation error and its propagation.

\subsection{Hybrid Numerical--Learning PDE Solvers}

Hybrid numerical--learning methods retain parts of the discrete structure or mechanistic model, while neural networks learn unknown dynamics, closure terms, or discretization errors. Interpretable spatiotemporal neural models have also been developed for PDE-governed distributed parameter systems \cite{ref44}. PDE information has further been incorporated as a physics-informed prior in Bayesian neural networks to obtain physically consistent forecasts from limited observations \cite{ref45}. Representative approaches include Universal Differential Equations, which combine known equations with learnable terms \cite{ref39}; methods that learn corrections to coarse-grid discretizations \cite{ref18,ref40}; and Solver-in-the-Loop \cite{ref20}, machine-learning-accelerated computational fluid dynamics \cite{ref19}, and Differentiable Turbulence \cite{ref41}, which incorporate differentiable solvers into training. These studies demonstrate that numerical structures can effectively constrain the learning process, but many rely on high-fidelity trajectories or are tailored to specific solvers. In contrast, the proposed method obtains the baseline evolution through low-cost numerical time stepping and learns an error correction from weak-form residuals without ground-truth trajectory supervision.

\section{PRIOR-GUIDED PHYSICS-CONSTRAINED EXTRAPOLATION WITHOUT GROUND-TRUTH TRAJECTORY SUPERVISION}

\subsection{Evolution-Operator Foundations and Methodological Motivation}

\begin{figure*}[!t]
\centering
\includegraphics[width=\textwidth]{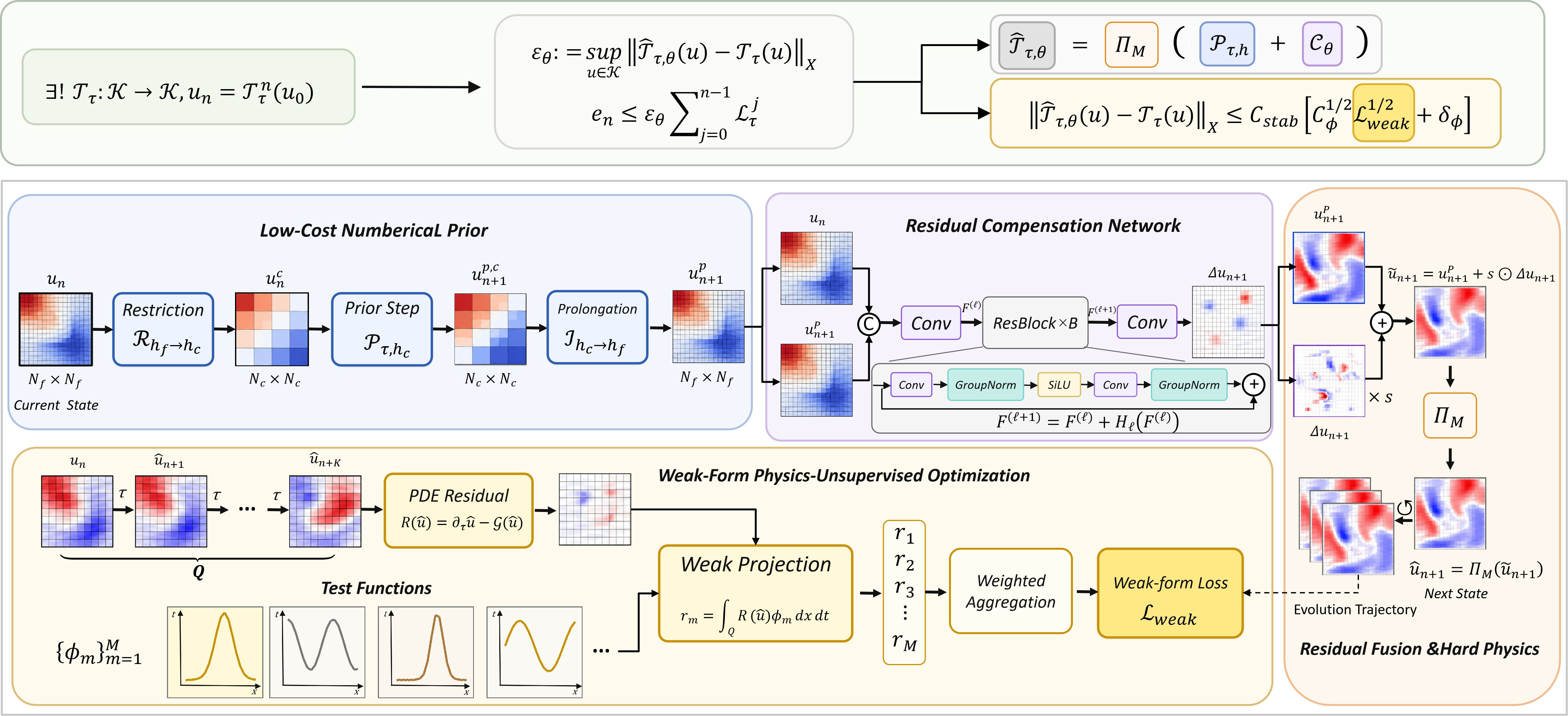}
\caption{Overall architecture and training workflow of the proposed method. The theoretical relations at the top connect recursive fixed-step evolution, one-step error propagation, and weak-form error constraints, with colored terms corresponding to their modules. Low-Cost Numerical Prior obtains the prior prediction through restriction, prior stepping, and prolongation. Residual Compensation Network concatenates the current state and prior prediction at \(C\) and outputs a correction. Residual Fusion \& Hard Physics scales the correction by \(s\), adds it to the prior prediction, and applies \(\Pi_{\mathcal M}\) to produce the next state for recursive rollout. Weak-Form Physics-Unsupervised Optimization constructs \(\mathcal L_{\mathrm{weak}}\) from the PDE residuals of short predicted trajectories through test-function projection and weighted aggregation, enabling training without ground-truth trajectory supervision.}
\label{fig:all_fig1}
\end{figure*}

As shown in Fig.~\ref{fig:all_fig1}, the proposed framework integrates a low-cost numerical prior, a residual compensation network, hard physical projection, and weak-form physics-unsupervised optimization for recursive long-time rollout.

This section starts from the temporal evolution law of autonomous PDEs and establishes a fixed-step evolution operator under the assumption that the corresponding initial-boundary value problem is well posed. It then analyzes the error between the exact evolution operator and its approximation, together with its propagation, thereby clarifying the roles of the numerical prior, neural correction, and weak-form constraints in improving operator approximation and long-time extrapolation.

\subsubsection{Fixed-Step Evolution Operator for Autonomous PDEs}

This work considers long-time extrapolation for autonomous PDEs. Let \(u(t)\) denote the system state. The governing equations are written in the general form

\begin{equation}
\partial_t u
=
\mathcal G(u;\mu),
\qquad
\mathcal B(u)=b,
\qquad
u(0)=u_0.
\tag{1}
\end{equation}

Here, \(\mathcal G\) denotes the evolution law defined by the PDE, \(\mu\) denotes the fixed physical parameters, and \(\mathcal B(u)=b\) represents time-invariant boundary conditions. Autonomy means that \(\mathcal G\) has no explicit dependence on absolute time: given the same current state, the system follows the same evolution over an equal subsequent time interval.

Assuming that the corresponding initial-boundary value problem is well posed, the exact evolution from the initial state to time \(t\) can be represented by the solution flow \(\mathscr S_t\). Autonomy further yields the semigroup relation

\begin{equation}
u(t)=\mathscr S_t(u_0),
\qquad
\mathscr S_{t+s}
=
\mathscr S_t\circ\mathscr S_s,
\qquad
\mathscr S_0=\mathcal I.
\tag{2}
\end{equation}

For a fixed time interval \(\tau\), define the exact one-step evolution operator as \(\mathscr T_\tau=\mathscr S_\tau\). Then,

\begin{equation}
u_{n+1}
=
\mathscr T_\tau(u_n),
\qquad
u_n
=
\mathscr T_\tau^{\,n}(u_0).
\tag{3}
\end{equation}

As shown in (3), exact long-time evolution can be generated by repeated composition of the same one-step operator. We therefore reformulate the learning target from the full spatiotemporal solution to a one-step evolution operator that can be applied recursively.

However, this result applies only to the exact operator and does not imply that its learned approximation \(\widehat{\mathscr T}_{\tau,\theta}\) also supports stable extrapolation. The approximate model introduces an error at each step, which continues to propagate under recursive application. Therefore, connecting the recursive structure of the exact operator to the long-time extrapolation capability of the model requires a relation between the one-step approximation error and its propagation. Building on this relation, the next subsection introduces the numerical prior, neural correction, and physics-based loss.

\subsubsection{Approximation of the Exact Evolution Operator and the Proposed Method}

The exact evolution operator can be repeatedly composed, but a practical model can only approximate it. Let \(\mathcal K\) denote the set of states reachable during extrapolation. We denote the learnable one-step evolution operator by \(\widehat{\mathscr T}_{\tau,\theta}\) and define its uniform one-step error as

\begin{equation}
\widehat u_{n+1}
=
\widehat{\mathscr T}_{\tau,\theta}(\widehat u_n),
\qquad
\varepsilon_\theta
=
\sup_{u\in\mathcal K}
\left\|
\widehat{\mathscr T}_{\tau,\theta}(u)
-
\mathscr T_\tau(u)
\right\|_X.
\tag{4}
\end{equation}

If the exact evolution operator has a local stability constant \(L_\tau\) on \(\mathcal K\), the predicted states remain in \(\mathcal K\), and both trajectories share the same initial state, then the error \(e_n=\|\widehat u_n-u_n\|_X\) at step \(n\) satisfies

\begin{equation}
e_{n+1}
\leq
L_\tau e_n+\varepsilon_\theta,
\qquad
e_n
\leq
\varepsilon_\theta
\sum_{j=0}^{n-1}L_\tau^j.
\tag{5}
\end{equation}

Thus, the recursive structure of the exact operator only provides the basis for extrapolation; long-time model accuracy depends on the one-step operator error and its propagation.

To reduce the difficulty of approximating the full operator \(\mathscr T_\tau\), we introduce a low-cost numerical prior \(\mathcal P_{\tau,h}\) and let the network learn the ideal correction to the prior error:

\begin{equation}
\begin{aligned}
\mathcal D_{\tau,h}(u)
&=
\mathscr T_\tau(u)-\mathcal P_{\tau,h}(u),
\qquad
\widehat{\mathscr T}_{\tau,\theta}(u)
\\
&=
\Pi_{\mathcal M}
\left[
\mathcal P_{\tau,h}(u)
+
\mathcal C_\theta
\left(u,\mathcal P_{\tau,h}(u)\right)
\right].
\end{aligned}
\tag{6}
\end{equation}

Here, \(\mathcal C_\theta\) approximates the ideal correction operator \(\mathcal D_{\tau,h}\), while \(\Pi_{\mathcal M}\) enforces the necessary physical constraints. The numerical prior provides a baseline approximation, and the convolutional neural network (CNN) learns only the correction to the prior error, thereby narrowing the operator-approximation range when the prior captures the dominant dynamics.



However, neither the exact operator nor the ideal correction is available during training, so Eq. (4) cannot be optimized directly. Both strong-form and weak-form residuals can provide physics-based criteria without ground truth; the weak form is adopted because stability estimates for evolution problems typically control state error through the dual norm of the spatiotemporal residual, which weak testing can discretize directly. If the PDE is stable and the test space is sufficiently rich,

\begin{equation}
\begin{aligned}
\left\|
v(\tau)-\mathscr T_\tau(u)
\right\|_X
&\leq
C_{\mathrm{stab}}
\left\|
\mathcal R(v;u)
\right\|_{\mathcal Y'}
\\
&\leq
C_{\mathrm{stab}}
\left(
C_\Phi^{1/2}
\mathcal L_{\mathrm{weak}}^{1/2}(v;u)
+
\delta_\Phi
\right).
\end{aligned}
\tag{7}
\end{equation}

Here, \(\mathcal Y'\) denotes the dual space of the spatiotemporal residual over one time step. The weak-form loss therefore converts the unavailable one-step operator error into a computable residual proxy that constrains how accurately \(\widehat{\mathscr T}_{\tau,\theta}\) approximates the one-step action of \(\mathscr T_\tau\). The conditions and derivation of Eq. (7) are provided in Appendix A.

During training, the same approximate operator is applied recursively over a short time window, while the weak-form residual constrains the entire predicted trajectory. Parameter sharing ensures that each time step follows the same evolution rule, and Eq. (5) shows that reducing the one-step approximation error correspondingly lowers its accumulated upper bound under recursive application. In summary, the numerical prior provides the baseline approximation, the CNN learns the prior correction, and the weak-form loss constrains the operator error without ground truth; together, these components form the proposed pathway to long-time extrapolation.



\subsection{Prior-Guided Residual Correction Operator}

This section presents the computational implementation of the approximate evolution operator \(\widehat{\mathscr T}_{\tau,\theta}\). Given the current state \(u_n\), a fixed numerical prior first produces a baseline prediction, after which the Residual CNN estimates a correction to the prior error. The next state is then obtained through residual fusion and the necessary physical constraints. A forward pass is summarized as

\begin{equation*}
u_n
\longrightarrow
u_{n+1}^{P}
\longrightarrow
\Delta u_{n+1}
\longrightarrow
\widehat u_{n+1}.
\end{equation*}

Neither the numerical prior nor the physical constraints contain trainable parameters; all network parameters reside in the correction operator \(\mathcal C_\theta\).

\subsubsection{Cross-Resolution Prior Embedding}

Let \(h_f\) and \(h_c\) denote the spatial scales of the target and prior grids, respectively, with \(h_c\geq h_f\). We represent the numerical prior as a fixed one-step time-marching operator \(\mathcal P_{\tau,h_c}\). The cross-resolution computation is

\begin{equation}
u_{n+1}^{P}
=
\mathcal I_{h_c\rightarrow h_f}
\left[
\mathcal P_{\tau,h_c}
\left(
\mathcal R_{h_f\rightarrow h_c}(u_n)
\right)
\right].
\tag{8}
\end{equation}

Here, \(\mathcal R_{h_f\rightarrow h_c}\) is the restriction operator from the target grid to the prior grid, and \(\mathcal I_{h_c\rightarrow h_f}\) is the prolongation operator. The prior module only needs to receive the current state, advance it by a fixed time step \(\tau\), and return a prediction containing the same physical variables as the original state; it is not tied to a specific numerical scheme.

\begin{figure}[!t]
\centering
\includegraphics[width=\columnwidth]{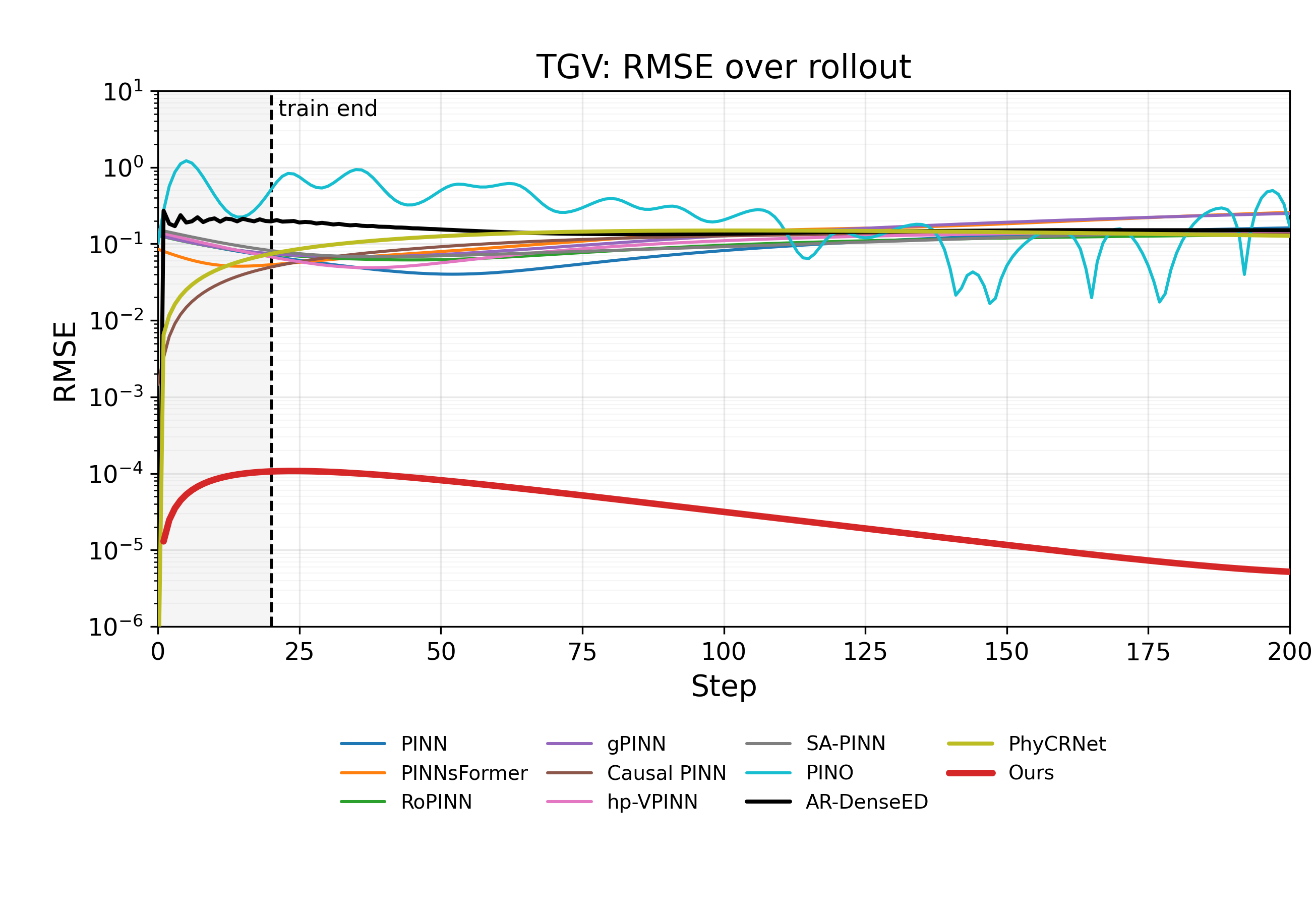}
\caption{Full-state RMSE for TGV over steps 0--200. Training ends at step 20.}
\label{fig:fig4-01a-tgv-rmse-ar-phycrnet-pic4}
\end{figure}

\begin{figure*}[!t]
\centering
\includegraphics[width=0.95\textwidth]{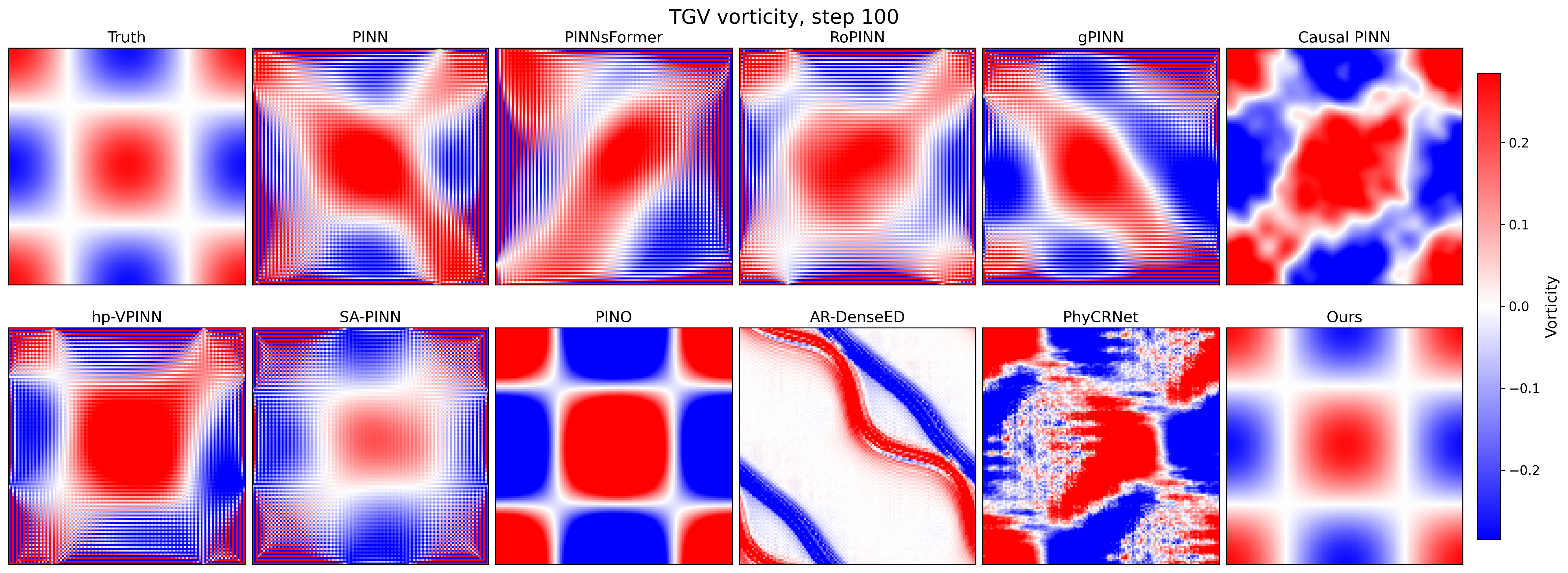}
\caption{TGV vorticity field at the representative extrapolation step 100. Because the analytical solution decays rapidly over time, the field at step 200 is provided in Fig. 19.}
\label{fig:fig4-01b-tgv-step100-ar-phycrnet-pic4}
\end{figure*}

\begin{figure}[!t]
\centering
\includegraphics[width= \columnwidth]{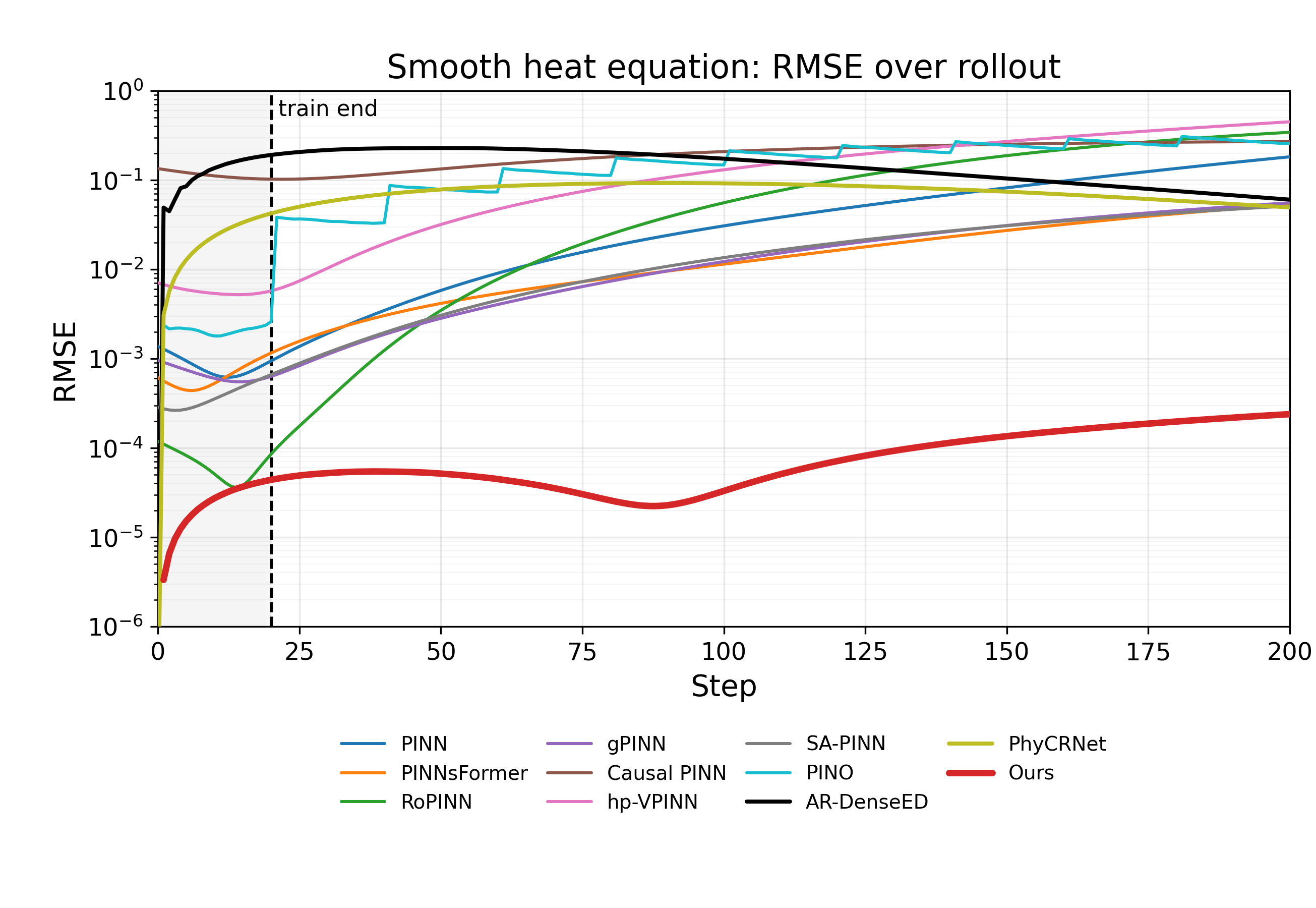}
\caption{RMSE for the smooth heat equation over steps 0--200. Training ends at step 20.}
\label{fig:fig4-02a-heat-rmse-ar-phycrnet-pic4}
\end{figure}

The prior is evaluated on a coarser grid to reduce the cost of baseline time stepping. If the number of nodes along each spatial direction decreases from \(N_f\) to \(N_c\), the prior degrees of freedom for a local grid-based method in \(d\) dimensions decrease to approximately \((N_c/N_f)^d\) of those on the target grid. This ratio applies only to the prior computation; prolongation and the CNN remain on the target grid. Errors introduced by coarse-grid discretization, restriction, and prolongation jointly constitute the prior error to be corrected by the network. The specific prior schemes, grid resolutions, and time steps are given in Appendix C and Appendix F-B.

\subsubsection{Prior-Error Correction with a Residual CNN}

The correction network approximates the error unresolved by the prior, and its input must contain the complete state required to determine the next-step evolution. We concatenate the current state and the upsampled prior prediction along the channel dimension and output a correction of the same dimensionality:

\begin{equation}
z_n
=
\operatorname{Concat}
\left(u_n,u_{n+1}^{P}\right),
\qquad
\Delta u_{n+1}
=
\mathcal C_\theta(z_n).
\tag{9}
\end{equation}

The network architecture must balance representation capacity, computational cost, and physical identifiability. Its effective receptive field should cover the dominant spatial scale of the prior error over one time step, without requiring global coupling among all locations. Because the network is repeatedly evaluated at every extrapolation step, we adopt a lightweight, moderate-capacity, fully convolutional architecture to limit both computational overhead and degrees of freedom left unconstrained by weak-form physical testing. Convolutional weights are shared over the full spatial field, with padding consistent with the boundary conditions. If tiling is necessary, overlapping regions are retained to avoid boundary artifacts from independent nonoverlapping patches.

A Residual CNN is adopted as the standard implementation. An initial \(3\times3\) convolution maps \(z_n\) to \(C\) hidden channels, followed by \(B\) residual blocks. Each block contains two \(3\times3\) convolutions, GroupNorm, and SiLU activations, and updates the features through a skip connection:

\begin{equation}
F^{(\ell+1)}
=
F^{(\ell)}
+
\mathcal H_\ell
\left(F^{(\ell)}\right),
\qquad
\ell=0,\ldots,B-1.
\tag{10}
\end{equation}

The network width \(C\), the number of residual blocks \(B\), and any required dilation rates jointly determine its capacity and receptive field; their values are selected according to the spatial scale of the prior error and the target resolution. The output convolution maps the hidden features to correction terms for the physical variables. Its weights and biases are initialized to zero, so at the start of training,

\begin{equation}
\mathcal C_{\theta_0}(z_n)=0,
\qquad
\widehat u_{n+1}
=
\Pi_{\mathcal M}
\left(u_{n+1}^{P}\right).
\tag{11}
\end{equation}

If the prior prediction already satisfies the physical constraints, the initial model is exactly equal to the prior. Zero initialization and residual output jointly limit the initial perturbation, allowing the network to correct the prior progressively rather than relearn the full evolution. The network size and boundary treatment are specified in Appendix F-A; temporal consistency is maintained by the parameter-shared one-step operator and the physics-based loss introduced in Section III-C.

\subsubsection{Residual Fusion and Physical Constraints}

The network output represents an additive correction to the prior prediction. To account for scale differences among physical variables, the correction is first scaled channel-wise and then fused with the prior prediction:

\begin{equation}
\widetilde u_{n+1}
=
u_{n+1}^{P}
+
s\odot\Delta u_{n+1}.
\tag{12}
\end{equation}

Here, \(s\) is the variable-scale vector, and \(\odot\) denotes channel-wise multiplication. The scales are determined by nondimensionalization or prescribed physical magnitudes and do not depend on ground-truth supervision.

The fused result is then mapped to the physically admissible set to obtain the final state:

\begin{equation}
\widehat u_{n+1}
=
\Pi_{\mathcal M}
\left(\widetilde u_{n+1}\right)
=
\widehat{\mathscr T}_{\tau,\theta}(u_n).
\tag{13}
\end{equation}

The operator \(\Pi_{\mathcal M}\) enforces the hard constraints required by the specific PDE, such as periodic or fixed boundary conditions, incompressibility, or variable positivity. If explicit projection is unnecessary for a given case, \(\Pi_{\mathcal M}=\mathcal I\). The numerical prior, CNN correction, and physical constraints thus constitute a complete one-step evolution. Parameter optimization and recursive application are introduced in the next section.

\begin{figure*}[!t]
\centering
\includegraphics[width=\textwidth]{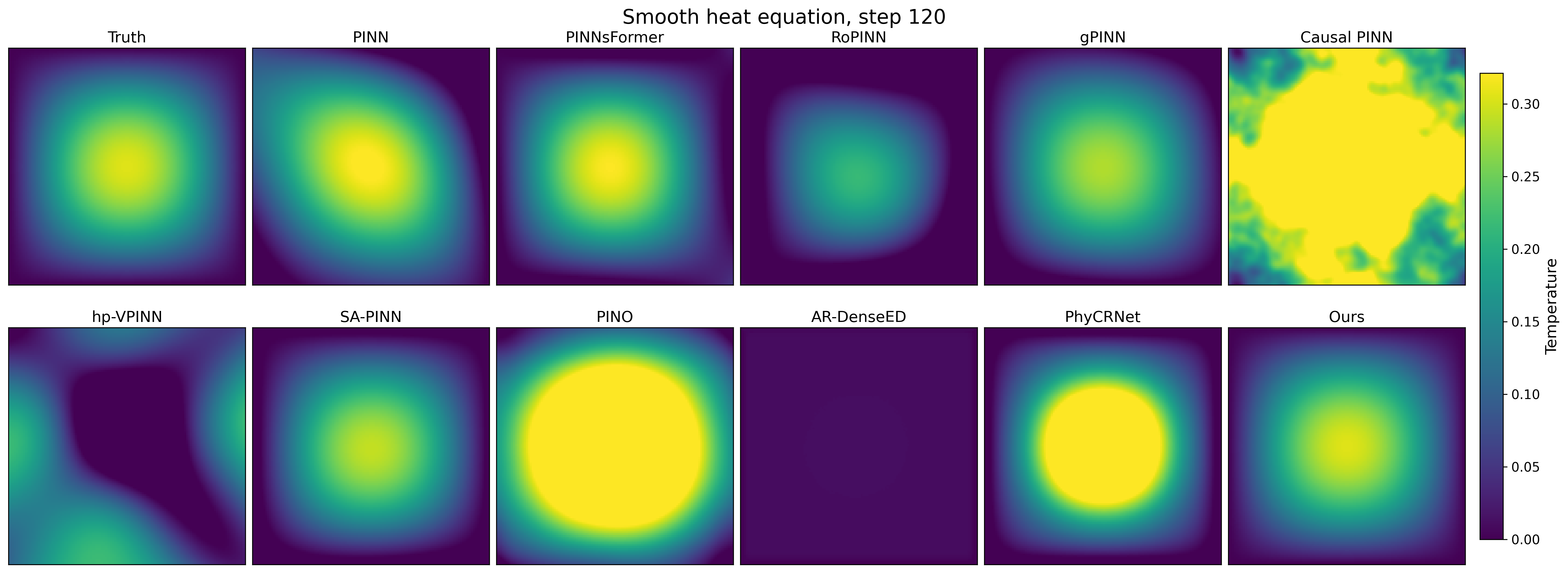}
\caption{Temperature field for the smooth heat equation at the representative extrapolation step 120. The field at step 200 is provided in Fig. 20.}
\label{fig:fig4-02b-heat-step120-ar-phycrnet-pic4}
\end{figure*}

\begin{figure}[!t]
\centering
\includegraphics[width=\columnwidth]{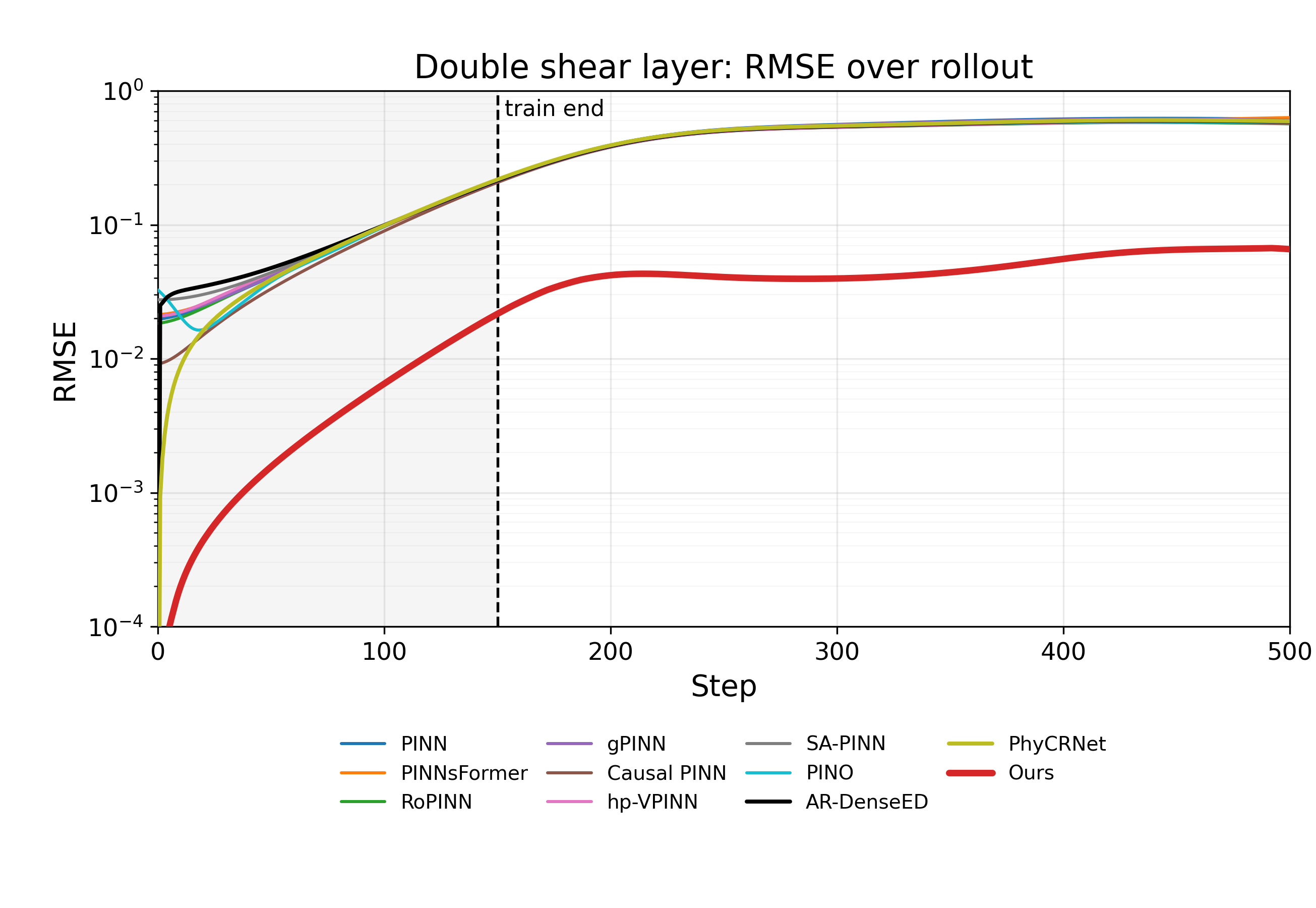}
\caption{Full-state RMSE for DSL over steps 0--500. Both the raw curve for Ours and its 15-step centered moving average are shown, with the latter included only for readability. All statistics are computed from the unsmoothed data.}
\label{fig:fig4-03a-dsl-rmse-ar-phycrnet-pic4}
\end{figure}

\begin{figure*}[!t]
\centering
\includegraphics[width=\textwidth]{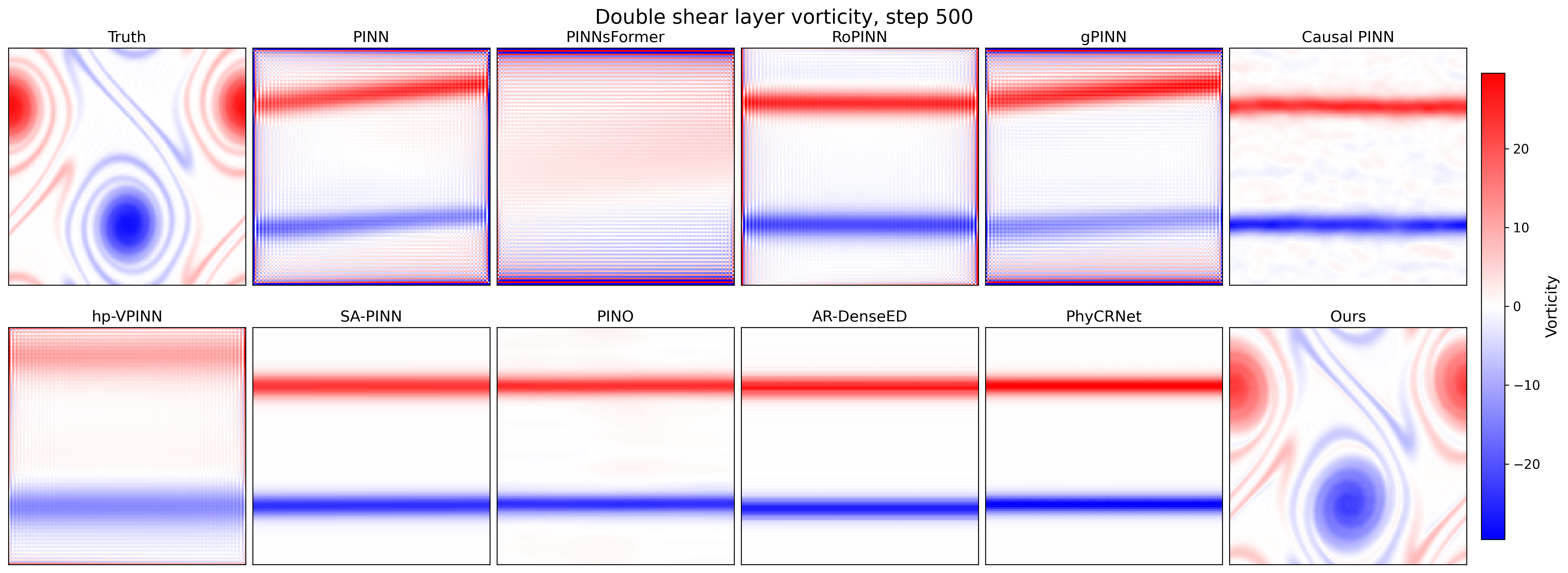}
\caption{DSL vorticity field at the final extrapolation step 500.}
\label{fig:fig4-03b-dsl-step500-ar-phycrnet-pic4}
\end{figure*}

\begin{figure}[!t]
\centering
\includegraphics[width=\columnwidth]{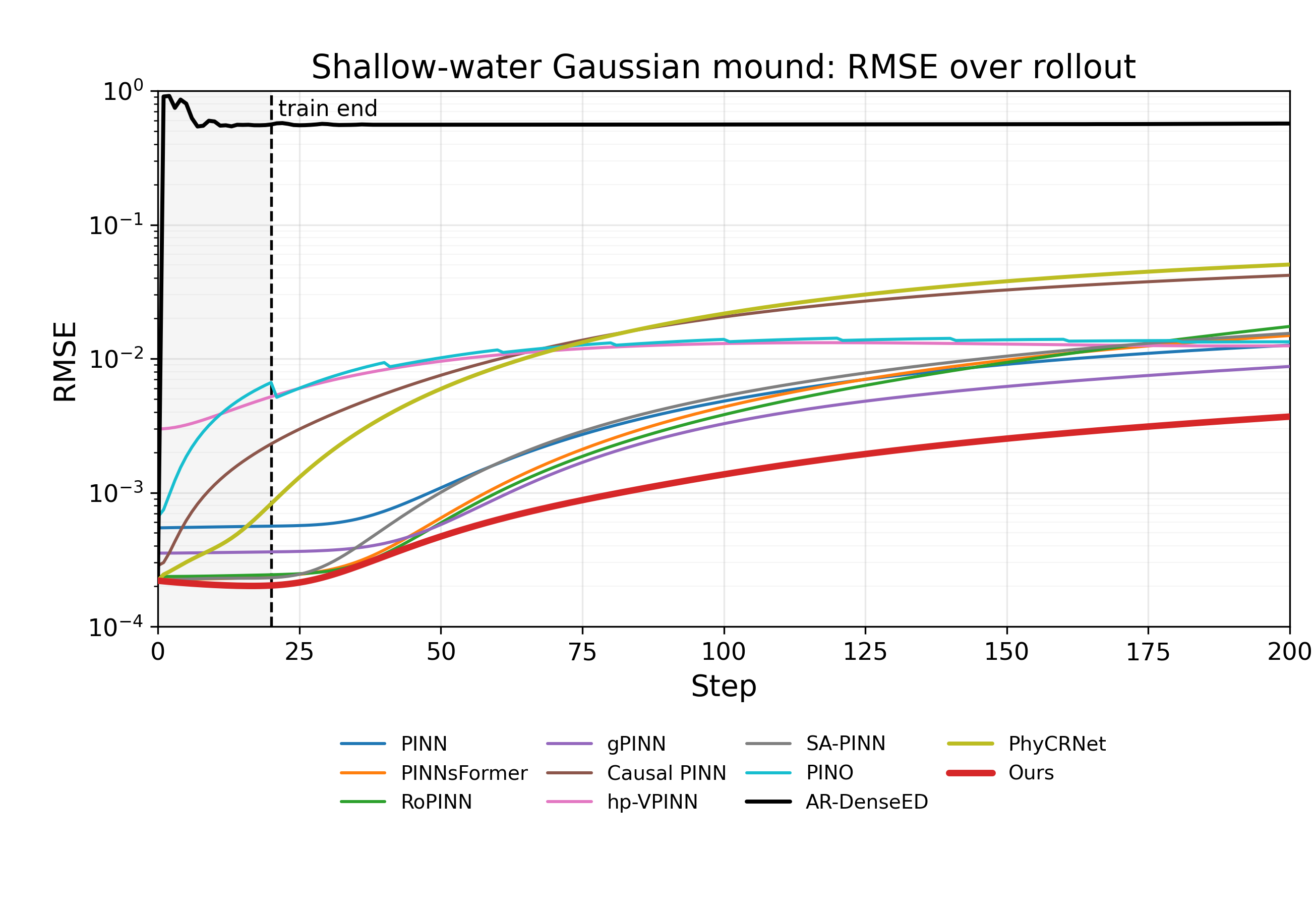}
\caption{Full-state RMSE for Gaussian Mound over steps 0--200.}
\label{fig:fig4-04a-gaussian-rmse-ar-phycrnet-pic4}
\end{figure}

\begin{figure*}[!t]
\centering
\includegraphics[width=\textwidth]{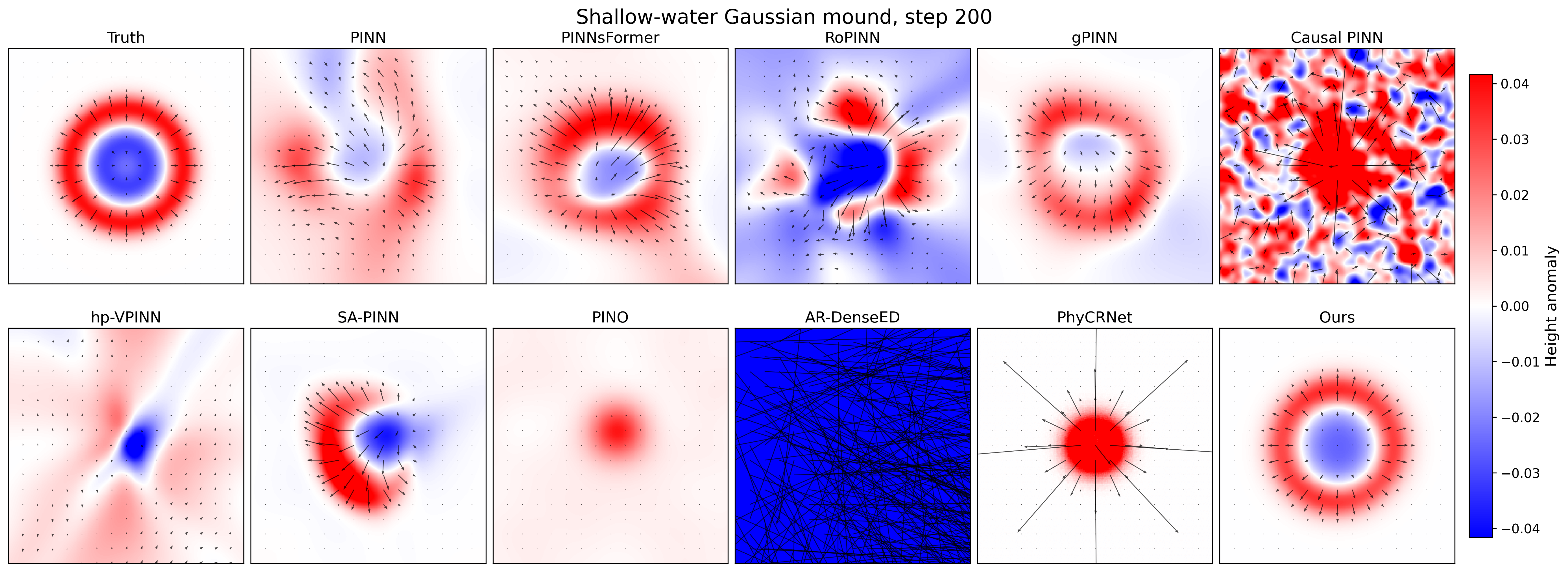}
\caption{Free-surface perturbation and velocity field for Gaussian Mound at step 200.}
\label{fig:fig4-04b-gaussian-step200-ar-phycrnet-pic4}
\end{figure*}

\subsection{Physics-Constrained Optimization without Ground-Truth Trajectory Supervision and Recursive Extrapolation}

This section describes how the composite operator is optimized without ground-truth trajectory supervision. The weak-form loss constrains short-window predictions to satisfy the governing equations, while necessary PDE-specific physical regularization further restricts inadmissible states. After training, the parameters are frozen, and the same operator is repeatedly applied for long-time extrapolation.

\subsubsection{Weak-Form Physics Constraint for One-Step Evolution}

Starting from state \(u_n\), the model generates a predicted sequence of length \(K\):

\begin{equation}
\widehat u_{n+k}
=
\widehat{\mathscr T}_{\tau,\theta}^{\,k}(u_n),
\qquad
k=1,\ldots,K.
\tag{14}
\end{equation}

Because the exact evolution operator is unavailable during training, the predicted sequence is evaluated through the governing equation. For the abstract equation \(\partial_tu=\mathcal G(u)\), define the residual as

\begin{equation}
\mathcal R(\widehat u)
=
\partial_t\widehat u-\mathcal G(\widehat u).
\tag{15}
\end{equation}

Select test functions \(\{\phi_m\}_{m=1}^{M}\) over the training spatiotemporal window \(Q\), and define

\begin{equation}
r_m
=
\int_Q
\mathcal R(\widehat u)\,\phi_m
\,\mathrm d\boldsymbol x\,\mathrm dt,
\qquad
\mathcal L_{\mathrm{weak}}
=
\frac{1}{M}
\sum_{m=1}^{M}
\omega_m|r_m|^2.
\tag{16}
\end{equation}

The conditions under which this loss serves as a proxy for the one-step error are given in Appendix A; its additional numerical and optimization advantages are discussed in Appendix B, and the test-function sets and quadrature rules used for each case are reported in Appendix F-C. Rigorous a posteriori error bounds for PDE-defined PINNs have likewise shown that computable prediction-error estimates can be obtained without access to the exact solution \cite{ref46}.

All initial and boundary conditions are enforced as hard constraints. The initial condition directly initializes the recursion, while the boundary conditions are preserved through boundary-consistent numerical operations and the physical projection \(\Pi_{\mathcal M}\). Therefore, no initial-condition or boundary-condition losses are added to Eq. (16).

\subsubsection{Overall Objective and Training without Ground-Truth Trajectory Supervision}

The overall training objective is

\begin{equation}
\mathcal L(\theta)
=
\mathcal L_{\mathrm{weak}}
+
\sum_{r=1}^{R}
\lambda_r\mathcal L_{\mathrm{phys}}^{(r)}.
\tag{17}
\end{equation}

Here, \(\mathcal L_{\mathrm{phys}}^{(r)}\) denotes PDE-specific physical regularization, such as constraints on divergence, conserved quantities, equilibrium states, or correction magnitude. These terms exclude nonphysical solutions that are not sufficiently constrained by the weak-form tests. Different physical variables are nondimensionalized or scale-normalized before evaluation to prevent variables with larger numerical magnitudes from dominating the optimization. Because the initial and boundary conditions are enforced by construction, the objective contains no initial-condition or boundary-condition losses and no ground-truth fitting terms.

Training is performed only over a prescribed short time window. For each window, the current model recursively generates a predicted sequence, after which Eq. (17) is evaluated and the parameters are updated. Model selection is based on physical residuals evaluated over held-out time windows and independent test functions. Reference numerical or analytical solutions are excluded from both the loss and model selection and are used only for error evaluation. The training windows, loss weights, and optimizer settings are given in Appendix F-D.

\subsubsection{Recursive Extrapolation with Frozen Parameters}

After training, the parameters \(\theta^*\) are frozen, and the same composite operator is repeatedly applied from the prescribed initial state:

\begin{equation}
\widehat u_0=u_0,
\qquad
\widehat u_{n+1}
=
\widehat{\mathscr T}_{\tau,\theta^*}(\widehat u_n),
\qquad
n=0,1,\ldots,N-1.
\tag{18}
\end{equation}

The parameters \(\theta^*\) are shared across all extrapolation steps. The state is not reset at the end of the training window, and neither ground-truth correction nor further network updates are introduced. Consequently, the one-step operator learned over the short training window is applied directly beyond the training interval, and its long-time performance depends entirely on one-step physical consistency and stability under recursive composition.
















\section{NUMERICAL EXPERIMENTS AND RESULTS}

This section evaluates the long-time extrapolation capability of the proposed method without ground-truth trajectory supervision using five PDE cases with distinct dynamics. It first presents the unified experimental setup and evaluation protocol, then compares the errors and field evolution of different methods for each case, and finally analyzes the sustained correction provided by the network relative to the numerical prior across all cases.

\subsection{Experimental Setup and Evaluation Protocol}

We select five benchmark cases with distinct dynamics to evaluate long-time extrapolation after training over a short time window. All methods roll out continuously from step 0 to the test endpoint. Table I summarizes the case settings and validation targets; the complete problem definitions, reference-solution sources, and numerical priors are provided in Appendix C, while the accuracy characterization and information-isolation protocol for the reference solutions are specified in Appendix C-F.

\begin{table*}[!t]

\caption{Benchmark Settings and Validation Targets\label{tab:table1}}

\centering

\footnotesize

\begin{tabularx}{\textwidth}{@{}YYYYY@{}}

\toprule

\textbf{Case} & \textbf{Governing equation} & \textbf{Training/test steps} & \textbf{Evaluation grid} & \textbf{Validation target} \\

\midrule

Taylor--Green vortex (TGV) & Incompressible Navier--Stokes equations & 0--20 / 0--200 & \(128^2\) & Smooth analytical flow \\

Smooth Heat Equation & Heat equation & 0--20 / 0--200 & \(128^2\) & Dissipation and near-zero stability \\

Double shear layer (DSL) & Incompressible Navier--Stokes equations & 0--150 / 0--500 & \(128^2\) & Nonlinear instability and vortex structures \\

Gaussian Mound & Shallow-water equations & 0--20 / 0--200 & \(256^2\) & Wave propagation and variable coupling \\

Cahn--Hilliard (CH) Two-Droplet & Cahn--Hilliard equation & 0--20 / 0--260 & \(256^2\) & Fourth-order conservation and interface evolution \\

\bottomrule

\end{tabularx}

\end{table*}

Training and test steps denote the physics-based optimization window and the full rollout interval, respectively.

We compare the proposed method with PINN \cite{ref9}, PINNsFormer \cite{ref32}, RoPINN \cite{ref33}, gPINN \cite{ref31}, Causal PINN \cite{ref13}, hp-VPINN \cite{ref23}, SA-PINN \cite{ref30}, PINO trained without ground-truth supervision \cite{ref24}, AR-DenseED \cite{ref14}, and PhyCRNet \cite{ref15}. AR-DenseED and PhyCRNet represent a physics-constrained autoregressive convolutional network and a convolutional recurrent PDE solver, respectively. Each method retains the core architecture and training mechanism of the original work and adopts the same governing equations, initial and boundary conditions, training window, and evaluation grid for each case. The adaptation and model-selection protocols are detailed in Appendix E.

All state variables are evaluated using per-step mean absolute error (MAE) and root mean square error (RMSE). The main text reports RMSE curves and representative extrapolated fields and analyzes the error reduction of the proposed method relative to the numerical prior in Section IV-C. Complete MAE results and longest-horizon fields are provided in Appendix D, the evaluation protocol in Appendix E, and the full implementation and randomness settings of the proposed method in Appendix F.

\subsection{Long-Time Extrapolation Results for Different Evolution Equations}

This section compares the long-time extrapolation performance of the proposed method and ten physics-informed learning methods across five benchmark cases. The main text reports per-step RMSE and representative extrapolated fields; complete problem definitions are provided in Appendix C, and the correction relative to the numerical prior is analyzed across all cases in Section IV-C. The vertical dashed lines in the curves indicate the end of training.

\subsubsection{TGV: Smooth Analytical Flow}

Fig. 2 presents the full-state RMSE over the rollout, and Fig. 3 shows the vorticity field at step 100.














This case evaluates extrapolation accuracy for a smooth analytical flow. Over steps 21--200, the average RMSEs of the proposed method and the best-performing baseline, PINN, are \(3.86\times10^{-5}\) and \(9.47\times10^{-2}\), respectively. At step 100, the proposed method retains a vortex structure consistent with the reference solution, whereas the competing methods exhibit clear deviations. These results show that the proposed method maintains high long-time accuracy for smooth evolution.

\subsubsection{Smooth Heat Equation: Linear Dissipative Evolution}

Fig. 4 presents the RMSE over the rollout, and Fig. 5 shows the temperature field at step 120.




















This case evaluates stability during long-time dissipation and the near-zero regime. Over steps 21--200, the average RMSEs of the proposed method and the best-performing baseline, PINNsFormer, are \(9.36\times10^{-5}\) and \(1.88\times10^{-2}\), respectively. At step 120, the proposed method preserves a smooth temperature distribution and the correct decay trend without evident rebound or nonphysical oscillations. These results show that the proposed method stably extrapolates linear dissipative evolution.

\subsubsection{DSL: Nonlinear Shear Instability}

Fig. 6 presents the full-state RMSE over the rollout, and Fig. 7 shows the vorticity field at step 500.




















This case evaluates error growth and vortex-structure preservation after nonlinear instability. Over steps 151--500, the average RMSEs of the proposed method and the best-performing baseline, Causal PINN, are \(4.77\times10^{-2}\) and \(5.04\times10^{-1}\), respectively. At step 500, the competing methods generally exhibit excessive smoothing or structural displacement, whereas the proposed method retains the primary vortex rolls and their surrounding structures. These results show that the proposed method better preserves spatial organization after nonlinear instability.

\subsubsection{Gaussian Mound: Smooth Hyperbolic Wave Propagation}

Fig. 8 presents the full-state RMSE over the rollout, and Fig. 9 shows the free-surface perturbation and velocity field at step 200.









This case evaluates the extrapolation of coupled variables during hyperbolic wave propagation. Over steps 21--200, the average RMSEs of the proposed method and the best-performing baseline, gPINN, are \(1.71\times10^{-3}\) and \(3.99\times10^{-3}\), respectively. At step 200, the proposed method accurately preserves the propagation position, symmetry, and velocity direction of the annular wave. These results show that the proposed method reduces multivariable coupling errors in smooth hyperbolic systems.

\subsubsection{CH Two-Droplet: Fourth-Order Conservative Phase-Field Evolution}

Fig. 10 presents the RMSE over the rollout, and Fig. 11 shows the phase-field distribution at step 200.







\begin{figure}[!t]

\centering

\includegraphics[width=\columnwidth]{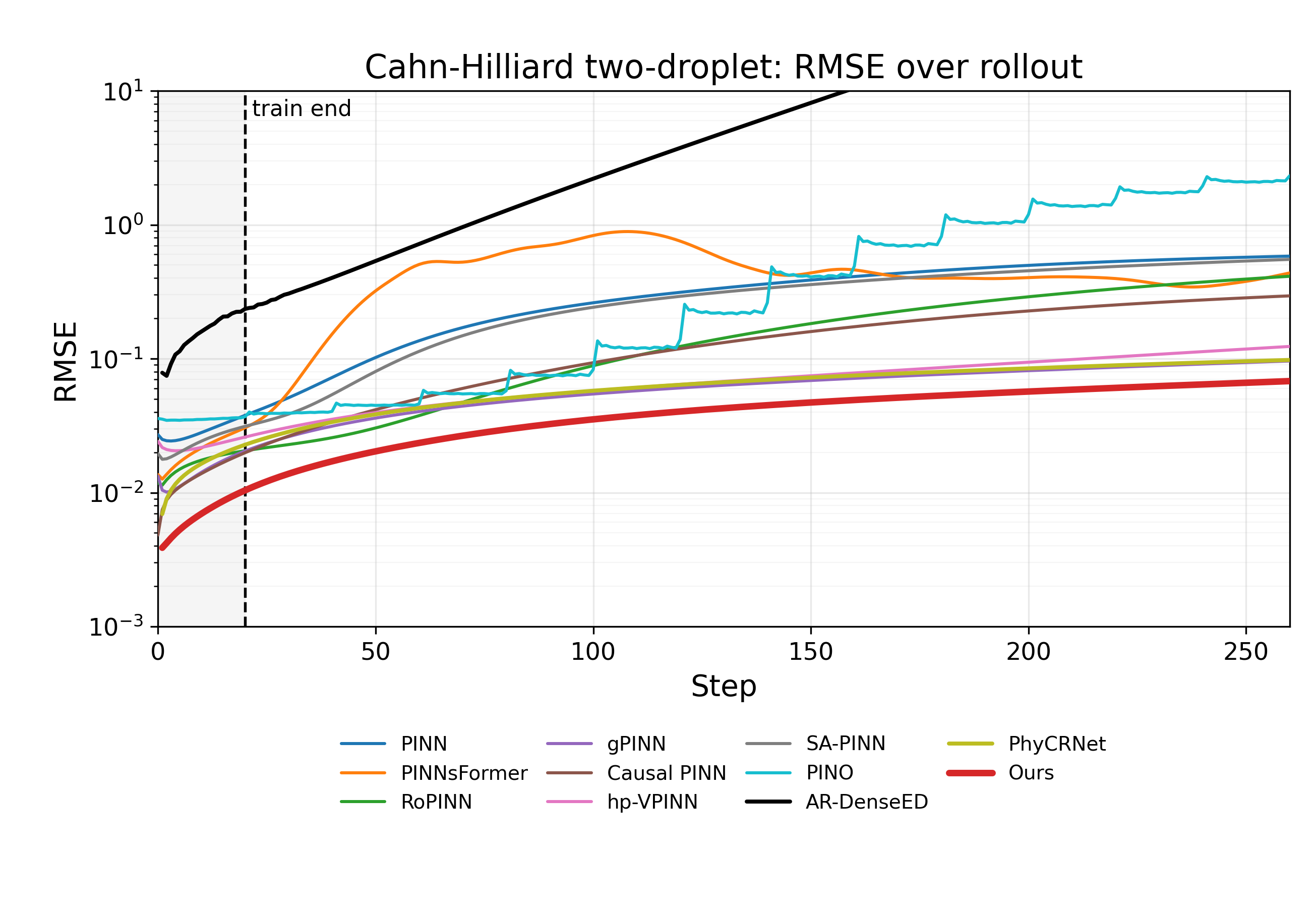}

\caption{RMSE for CH Two-Droplet over steps 0--260.}

\label{fig:fig4-05a-ch-rmse-corrected-no-inset-pic5}

\end{figure}

\begin{figure*}[!t]

\centering

\includegraphics[width=\textwidth]{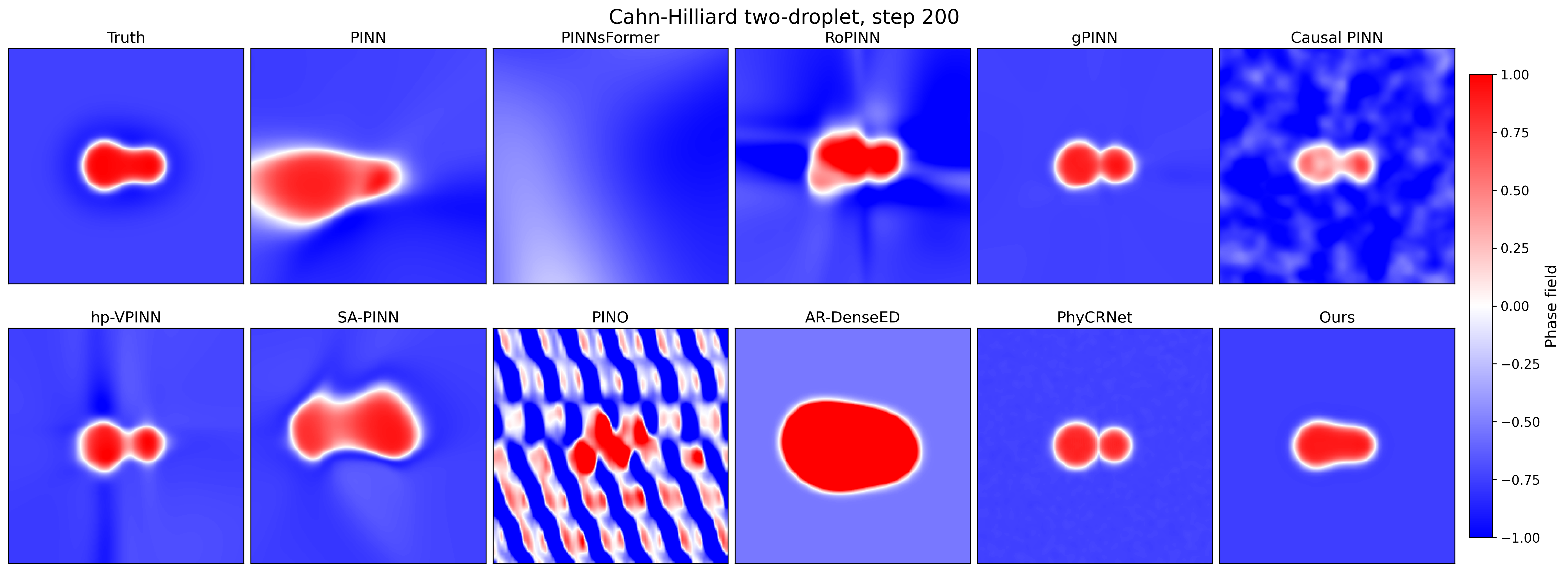}

\caption{Phase-field distribution for CH Two-Droplet at the representative extrapolation step 200.}

\label{fig:fig4-05b-ch-step200-corrected-pic5}

\end{figure*}

This case evaluates interface and topology evolution in a fourth-order conservative system. Over steps 21--260, the average RMSEs of the proposed method and the best-performing baseline, gPINN, are \(4.29\times10^{-2}\) and \(6.40\times10^{-2}\), respectively. At step 200, the proposed method preserves clear phase-separation interfaces and the main droplet structures. These results show that the proposed method applies not only to second-order transport and diffusion problems but also to fourth-order conservative phase-field evolution.

\subsection{Cross-Case Analysis of Prior Correction}

To isolate the contribution of the learned correction in the full model, we compare its per-step error with that of the numerical prior and examine whether it consistently improves recursive predictions beyond the training interval.

For each case, the full model and the numerical prior are rolled out continuously from step 0, and their RMSEs are computed. Let \(E_{c,m}^{(n)}\) denote the RMSE of method \(m\) for case \(c\) at step \(n\), where \(m\in\{\mathrm{prior},\mathrm{full}\}\) denotes the numerical prior and the full model, respectively. The normalized rollout progress and peak-normalized error reduction are defined as

\begin{equation}
s_c^{(n)}
=
\frac{n}{N_c},
\qquad
I_c^{(n)}
=
100\%
\frac{
E_{c,\mathrm{prior}}^{(n)}
-
E_{c,\mathrm{full}}^{(n)}
}{
\displaystyle
\max_{0\leq j\leq N_c}
E_{c,\mathrm{prior}}^{(j)}
}.
\tag{19}
\end{equation}

Here, \(N_c\) is the final test step, and \(I_c^{(n)}>0\) indicates that the full model has a lower error than the numerical prior. Normalization by the peak prior RMSE aligns the error scales across cases and avoids excessive amplification of pointwise relative ratios when the errors approach zero.







\begin{figure}[!t]

\centering

\includegraphics[width=\columnwidth]{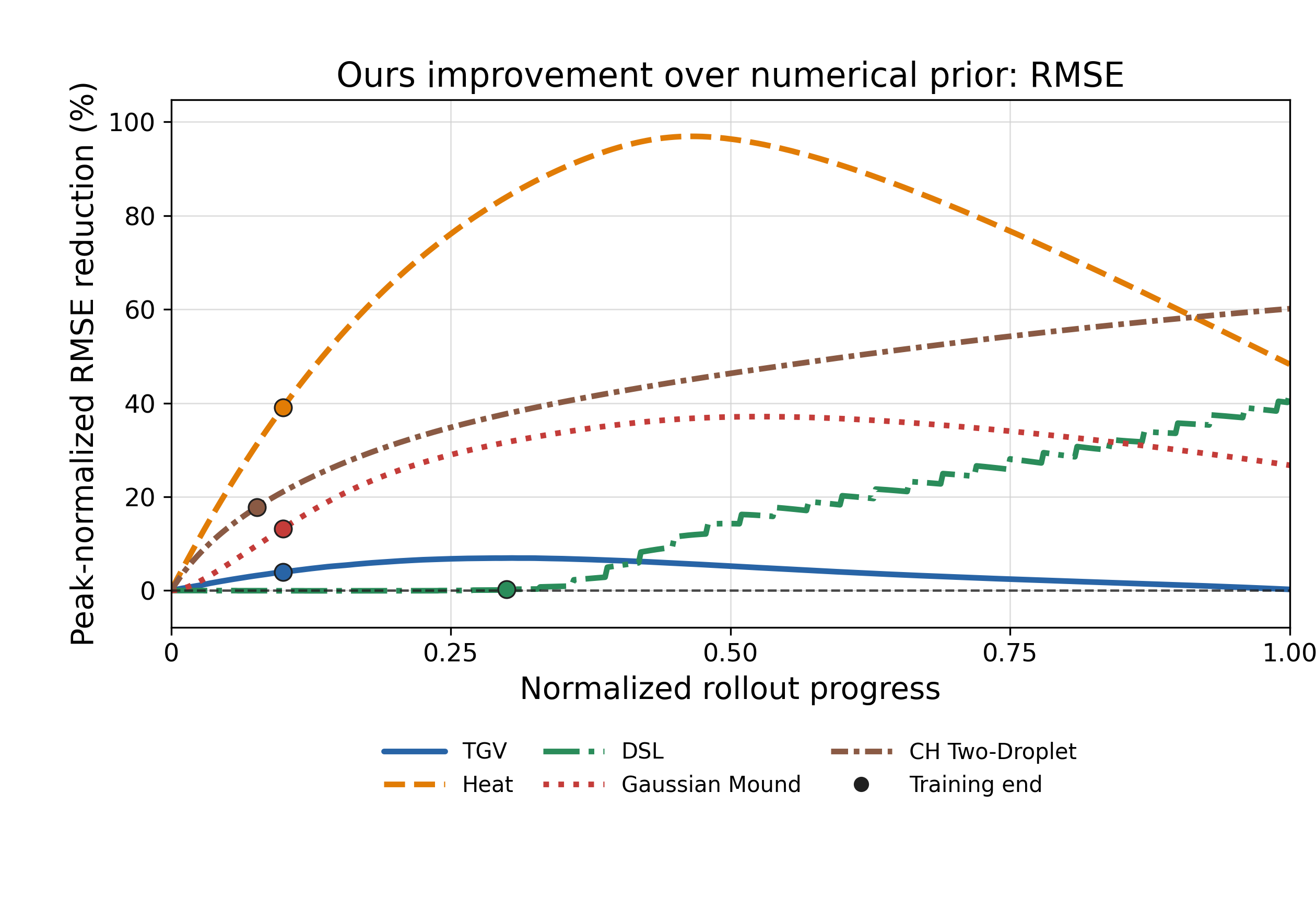}

\caption{Per-step RMSE reduction of the proposed method relative to the numerical prior. The circles mark the end of training for each case, with the training interval on the left and the extrapolation interval on the right; positive values indicate that Ours outperforms the numerical prior. The corresponding MAE results are provided in Appendix D.}

\label{fig:fig4-06-rmse-improvement-ts13-circle-trainmark}

\end{figure}

As shown in Fig. 12, all curves start from zero gain at the common initial state, and all five cases maintain positive gains at every extrapolation step. TGV and Heat are dominated by solution-amplitude decay, and their gains gradually decrease in the middle and late stages. DSL and CH are dominated by vortex-structure and interface evolution over the test interval, and their gains generally increase during rollout. Gaussian Mound maintains a relatively stable positive gain during wave propagation.

These differences show that the correction magnitude does not vary monotonically with equation complexity but depends on the accuracy of the numerical prior and how its error accumulates during evolution. The MAE results in Appendix D follow the same trend as the RMSE results, showing that the learned correction consistently reduces the prior error over the extrapolation intervals of different PDEs and that the performance improvement does not arise solely from the numerical prior.

\section{CONCLUSION}

This study investigates long-time extrapolation without ground-truth trajectory supervision for deep learning-based PDE solvers. To address the dependence of neural operators on trajectory data and the tendency of physics-informed methods to accumulate errors during recursive extrapolation, we start from the existence of the discrete evolution operator for an autonomous PDE and the propagation of its approximation error, interpret long-time rollout as repeated approximation of the same exact evolution mapping, and characterize extrapolation stability in terms of controlling the one-step approximation error and its recursive propagation.

Based on this understanding, we develop a numerical-prior-guided, physics-constrained framework for extrapolation without ground-truth trajectory supervision. To reduce the difficulty of approximating the full evolution operator, we introduce a low-cost numerical prior to provide the baseline evolution; to correct the error unresolved by the prior, we further construct a correction network; and to constrain the one-step error without ground-truth trajectory supervision, we employ the weak-form PDE residual as a computable proxy for its upper bound, following Eq. (7). Consequently, after training, the model can be applied recursively from the initial state, yielding a unified evolution approximation for long-time extrapolation.

The numerical experiments include five benchmark cases---TGV, the smooth heat equation, DSL, Gaussian Mound, and Cahn--Hilliard Two-Droplet---spanning smooth analytical flow, linear dissipation, nonlinear shear instability, hyperbolic wave propagation, and fourth-order conservative phase-field evolution. The proposed method outperforms the corresponding numerical prior in every case and achieves more stable long-time extrapolation than multiple physics-informed learning baselines. A cross-case analysis of prior correction further shows that the learned correction consistently reduces the prior error throughout the extrapolation interval, indicating that the performance gains do not arise solely from the numerical prior. The primary role of the method is not to replace numerical evolution rules, but to build on an existing physical prior to construct a more tractable and tightly constrained approximation of the evolution operator, thereby mitigating error accumulation during long-time rollout.

This study has several limitations. Current experiments focus mainly on two-dimensional regular grids with fixed initial and boundary conditions, and applicability to complex geometries, multiphysics coupling, high-dimensional systems, and parametric PDE families requires further validation. Moreover, the test-function set currently requires PDE-specific adjustment; future work will investigate automatic selection algorithms and their effects on the test-space error term in Eq. (7). Future work will also analyze the relationship between numerical-prior accuracy and long-time error propagation and extend the framework to more complex scientific computing settings.

\end{document}